\documentclass{article}
\usepackage{spconf,amsmath,graphicx}
\usepackage[colorlinks,linkcolor=red]{hyperref}

\title{Difftalker: Co-Driven Audio-Image Diffusion for Talking Faces via Intermediate Landmarks}
\name{Zipeng Qi$^{1,2\dagger}$, Xulong Zhang$^{1\dagger}$, Ning Cheng$^{1\ast}$\thanks{$^\dagger$These authors have equal contributions.\\$^{\ast}$Corresponding author: Ning Cheng (chengning211@pingan.com.cn)}, Jing Xiao$^{1}$, Jianzong Wang$^{1}$}
\address{$^1$Ping An Technology (Shenzhen) Co., Ltd.\\
        $^2$Beihang University
        }

\begin{document}
%
\maketitle
\begin{abstract}

Generating realistic talking faces is a complex and widely discussed task with numerous applications. In this paper, we present DiffTalker, a novel model designed to generate lifelike talking faces through audio and landmark co-driving. DiffTalker addresses the challenges associated with directly applying diffusion models to audio control, which are traditionally trained on text-image pairs. DiffTalker consists of two agent networks: a transformer-based landmarks completion network for geometric accuracy and a diffusion-based face generation network for texture details. Landmarks play a pivotal role in establishing a seamless connection between the audio and image domains, facilitating the incorporation of knowledge from pre-trained diffusion models. This innovative approach efficiently produces articulate-speaking faces. Experimental results showcase DiffTalker's superior performance in producing clear and geometrically accurate talking faces, all without the need for additional alignment between audio and image features.

\end{abstract}
\section{Introduction}
\label{sec:intro}
The field of talking face synthesis has attracted significant attention for its capacity to produce facial images that precisely synchronize with input audio, with a particular emphasis on depicting the mouth's shape. This task holds immense potential across various domains, including digital humans~\cite{thite2022digital}, and virtual video conferences~\cite{ji2022eamm}.

The most challenging aspect is to maintain the identifying information of the speaker while generating a realistic facial structure.  In recent years, several GAN-based methods~\cite{chen2019hierarchical, song2018talking, yin2022styleheat} have emerged, integrating audio and facial features within a generative network while employing a discriminator network to assess result authenticity. Wave2Lip~\cite{prajwal2020lip} introduces a distinct set of complete reference face images to retain the speaker's identity and additionally employs a weak lip-sync discriminator to enhance result accuracy. StyleSync~\cite{guan2023stylesync} efficiently leverages identified information encoded within a $w^+$ space and audio features to generate an accurate lower half of the facial image. Other works, such as MakeitTalk~\cite{zhou2020makelttalk}, predict landmarks by utilizing input speech features to manipulate facial geometry, subsequently enhancing facial textures through an additional image translation network. However, GAN-based methods require the concurrent training of both generator and discriminator networks, typically demanding large and comprehensive datasets.

Diffusion-based generative models~\cite{rombach2022high,hu2021lora,zhang2023adding} have garnered significant attention due to their remarkable generative capabilities. These models employ a progressive approach wherein Gaussian noise is incrementally introduced into an image. Subsequently, an attention-Unet is employed to iteratively reconstruct the clear image. DAETalker~\cite{du2023dae} encodes image features as controllable information and trains an aligner between audio and image features. In the study conducted by Zhu~\cite{zhua2023audio}, a concatenation approach is utilized to merge aligned personalized attribute features and synchronized audio features, which are then fed into a diffusion-based decoder for generating face images. The diffusion-based model effectively leverages a pre-trained CLIP model to align image features with text features, enabling text features to serve as effective conditioning for the generation process. Nevertheless, aligning audio and image features remains a challenge for diffusion models.

To tackle the aforementioned challenges, we propose leveraging landmarks as an intermediary outcome, effectively bridging the gap between these two domains. In this paper, we present DiffTalker, a framework that encompasses two agent networks: a transformer-based landmarks completion network for ensuring geometric accuracy, and a diffusion-based face generation network for enhancing texture details. Our input comprises three key components: audio, an upper-face image, and the corresponding upper landmarks sourced from~\cite{wang2021one}. The output includes the completed landmarks and the entire facial representation. For additional insights, please refer to our project details on our website: \href{https://qizipeng.github.io/DiffTalker/}{project}. 

The primary contributions of this study can be succinctly summarized as follows:
(1) We introduce DiffTalker, a novel diffusion-based framework for generating talking faces guided by both audio and landmarks. (2) We leverage landmarks to establish a vital link between audio and image features. We design two agent tasks: audio-driven landmarks completion and landmark-driven face generation.

\begin{figure*}[htbp]
\centering
\includegraphics[width=\linewidth]{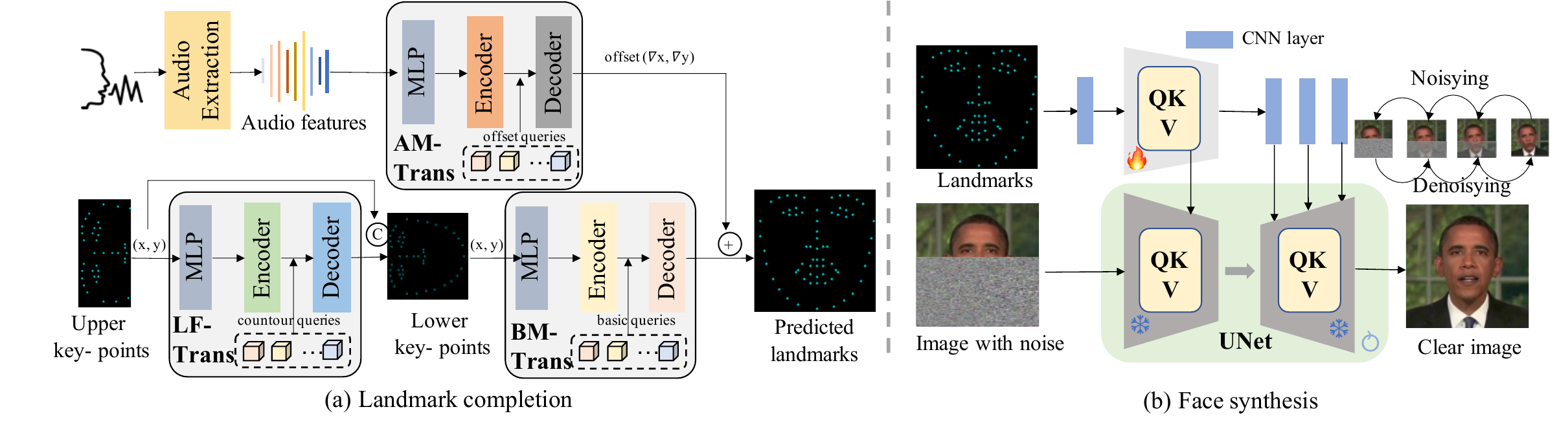}
\caption{We propose DiffFace, a method for generating a face using audio guidance. DiffFace comprises two proxy sub-networks: a landmark completion network and a face texture generation network. The first network takes the upper face's key points as input and outputs the lower key points, which form the mouth shape based on the additional audio feature input. The second network is based on the diffusion model, which incorporates the predicted landmarks as additional information to ensure accurate facial geometry in the generated results.}
\label{overview}
\end{figure*}

\section{method}
\label{method}
\subsection{overview}
As illustrated in Figure~\ref{overview}, we decompose DiffTalker into two agent networks: (1) an audio-guided landmark prediction network and (2) a landmark-guided face generation network. The landmark prediction network takes the upper half of the landmarks as input and generates the remaining key points with the support of audio features. It comprises three transformer-based modules, responsible for predicting the lower half of the face (\textbf{LF-Trans}), establishing the basic mouth shape (\textbf{BM-Trans}), and adjusting the mouth based on audio input (\textbf{AM-Trans}). We view the landmark results as supplementary information that assists the diffusion model in generating precise facial geometry, particularly for mouth shape. The diffusion-based face prediction network takes both Gaussian noise and the predicted landmarks as input, producing the facial texture. Similar to most methods~\cite{zhang2023adding, rombach2022high}, the landmark features interact with image features through cross-attention. By employing landmarks, we bridge the domain gap between the audio and image spaces without the need for training an additional aligner network.

\subsection{Audio Features Extraction}
To efficiently extract audio features, in line with prior audio-driven methods~\cite{cudeiro2019capture}, we employ the widely-recognized DeepSpeech~\cite{amodei2016deep} model, which predicts a 29-dimensional feature. Additionally, we incorporate a temporal convolutional network, akin to AD-NeRF~\cite{guo2021ad}, to simultaneously consider multiple consecutive frames of audio features. This network generates $a\in R^{16\times 29}$ based on the sixteen neighboring frames. The audio feature extraction network is optimized concurrently with the landmark completion network below.

\subsection{Audio-driven Landmark Completion}

The first subnetwork is the landmark completion network. Given the audio features $a$ and the upper half of the landmarks predicted by~\cite{wang2021one}, the landmark completion network generates results in two parts. These outcomes encompass the key points of the lower half of the facial contour and the mouth points that correspond to the input audio features. In this study, we assume that each landmark consists of 68 key points. The audio information primarily influences the distribution of the mouth points while having minimal impact on other facial points. To integrate the influence of audio features exclusively on the mouth key points, we employ different transformers. We selected the transformer as the foundational architecture due to its suitability for processing sequence data.

Firstly, LF-Trans utilizes the upper half of the landmarks to predict the distribution of contour key points in the lower half of the face. Specifically, we represent the input vector of each key point as its normalized coordinate $(x, y)$ and extract features using a coordinate MLP. LF-Trans incorporates self-attention mechanisms to effectively capture long-range interactive relationships between the coordinate features. The resulting features are subsequently fed into a cross-attention module, which utilizes a set of adaptable features as contour queries. The number of learnable queries equals the number of lower contour key points. Following the cross-attention, another coordinate MLP is employed to predict the coordinate values of the lower contour points. The fundamental contour points of the face consist of the input key points and predicted contour points. Moreover, BM-Trans utilizes these fundamental contour points to estimate the initial mouth coordinates, which are influenced by the arrangement of other points such as the eyes and head pose. AM-Trans takes audio features $a$ as input to predict the offset of each fundamental mouth point, primarily affecting the specific details of mouth shapes. By employing separate transformers for points in different regions, we can specify that the audio features only impact the shape details of the mouth, while the fundamental position of the mouth is determined in conjunction with other key points.

We optimize the weights of three transformer networks and audio extraction network by minimizing the mean of the square difference of the coordination of all predicted points and the ground truth:
\begin{equation}
\mathcal{L} = \frac{1}{N} \sum \limits_{p \in R}\Big[\Vert {C(p)'}-C(p)\Vert_2^2\Big]
\end{equation}
where the $R$ is the point set to be predicted, $C(p)'$ is the predicted results, $C(P)$ is the ground truth and N is the number of $R$.

\subsection{Landmark-driven Face Synthesis}
The second sub-agent is a diffusion-based face synthesis network. It takes the predicted landmarks and the upper face as input and produces the complete face texture. The face generation network comprises two branches, as depicted in Fig.~\ref{overview}. The upper branch extracts landmark features to incorporate geometry information, while the lower branch employs a frozen Unet to extract upper facial features as identity information guidance. The frozen Unet in the second branch effectively provides priors that have been trained on a large dataset.

The adding noise process in general diffusion is as follows:
\begin{equation}
Z_t = \sqrt{\overline{\alpha_t}}Z_0 + \sqrt{1-\overline{\alpha_t}}\epsilon_t
\end{equation}
where the $Z_0$ is the original image features from a pre-trained VAE~\cite{kingma2013auto}, $\overline{\alpha_t}$ is the hyper-parameter, and $\epsilon$ is a random Gaussian noise. $Z_t$ is the noised feature in the $t\in [1,n]$ step. The goal of the diffusion models is to learn the $\epsilon_t$ by a conditional Unet as follows:
\begin{equation}
\epsilon_{t}^{'} =\text{Unet}(Z_t, t, y)
\end{equation}
where the $y$ is the guided information. 

In our work, we use the upper facial features as a reference and modify the process of introducing noise. This approach enables us to establish distinct noise scales for the upper and lower regions:
 \begin{equation}
Z_t =\left \{
      \begin{array}{cl}
      Z_0 \quad i<=h//2\\
      \sqrt{\overline{\alpha_t}}Z_0 + \sqrt{1-\overline{\alpha_t}}\epsilon \quad i>h//2 \\
      \end{array}
      \right.
\end{equation}
The advantages of this way are that: (1) the clear image features can efficiently guide the denoising direction and provide the identifying information; (2) Generating realistic results for certain areas, such as the eyes, can be challenging. However, it is efficient in our way to generate images with more real details. To optimize the parameters, we use the $\epsilon_{t}$ as the ground truth and calculate the $L2$ error between the estimated noise and the ground truth:
\begin{equation}
loss_t = \| \epsilon_t - \epsilon_{t}^{'}\| 
\end{equation}
It should be noted that we exclusively compute the loss for the lower area. This is because we utilize the original image features to complete the upper area during the $0_{th}$ sampling step.

To ensure the precision of the generated facial geometry, we incorporate the predicted landmarks as guiding geometry information. In particular, we employ a dual-branch architecture, as illustrated in Fig.~\ref{overview}. The lower branch is a pre-trained Unet with fixed parameters, drawing upon prior knowledge from a vast text image dataset. The upper branch comprises an encoder, akin to the pre-trained Unet, and additional convolutional layers for extracting features from the completed landmarks. The encoder's landmark features interact with the pre-trained Unet decoder through a cross-attention mechanism. This integration allows the model to infuse geometry information from the landmarks into the original generation process. It is important to note that we update the parameters of the encoder in the upper branch to enable the network to adapt to the new data.

In the testing stage, the diffusion models generate the corresponding images from a random Gaussian noise $N(\textbf{0}, \textbf{1})$. 
The denoising process includes t-loop steps. Given the estimated $\epsilon_t$, we can use the forward formulation to calculate the result of single-step denoising:
\begin{equation}
z_{t-1} =\sqrt{\overline{\alpha_{t-1}}} \frac{z_t -\sqrt{1-\alpha_t}\epsilon_{t}^{'}}{\sqrt{\overline{\alpha_t}}}+\sqrt{1-\overline{\alpha{t-1}} } \epsilon_{t}^{'}
\end{equation}

\section{Experiments}
In this section, we evaluate our method's performance and compare it with several baselines, including Wav2Lip~\cite{prajwal2020lip}, DAE-Talker~\cite{du2023dae}, and MakeitTalk~\cite{zhou2020makelttalk}.£
\subsection{Dataset}
In this study, the Obama's address dataset~\cite{suwajanakorn2017synthesizing} is employed for both training and testing purposes. A short video sequence accompanied by an audio track is collected. Approximately 7,000 frames are used for training, while the remaining 1,000 frames are reserved for testing. To ensure compatibility with the input size of the VAE, all input frames are resized to $512 \times 512$ pixels. Additionally, a landmark prediction network~\cite{wang2021one} is utilized to generate the coordinates of upper key points corresponding to each frame.

\subsection{Implementation Details}
We implemented our framework using PyTorch. Both of the two agent networks were trained on a single 3090 GPU using the Adam solver, with initial learning rates of 0.0005 and 0.0001, respectively. The face synthesis network employs version 1.5 of the basic diffusion model.


\subsection{Landmark completion}
\begin{figure}[htbp]
\centering
\includegraphics[width=\linewidth]{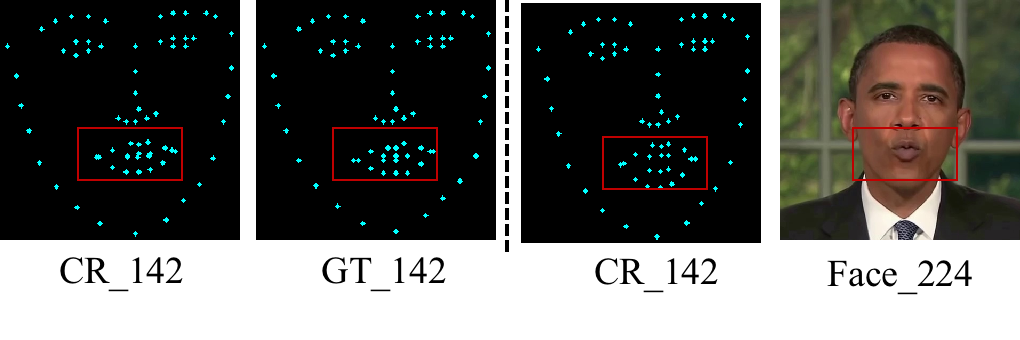}
\caption{The results of the landmark completion. $CR_n$: completion results of $n_{th}$ frame; $GT_n$: the ground truth of $n_{th}$ frame; $Face_n$: the face image of $n_{th}$ frame.}
\label{landmark}
\end{figure}

Figure~\ref{landmark} presents a comparison between landmark completion and ground truth. In the left portion of the results, it is evident that the landmark completion network accurately predicts the coordinates of the lower key points, which provide geometry information for the subsequent face synthesis network. To assess the impact of audio features on mouth key points, we use mismatched landmarks and speech features as inputs, such as employing the upper landmark of the $142_{th}$ frame and the audio features of the $224_{th}$ frame. The third column of the results demonstrates that the shape of the mouth accurately aligns with the facial image, confirming the success of our design by solely utilizing the offset of the mouth key points.

We assess the quantitative results of our landmark completion network and find that the average Euclidean distance for all the normalized testing results is 0.002. Additionally, we modified by removing the AM-Trans and incorporating audio features into the BM-Trans, which directly predict the coordinates of the mouth. Consequently, the average Euclidean distance increased to 0.005.

\subsection{Face Synthesis}
To assess the performance of the generated faces, we utilized Five metrics: LSE-D, LSE-C, LD (The landmark distance between predictions and GTs), PSNR, and SSIM. From Table~\ref{compaire}, we can find that our method ($Ours^2$) outperforms the GAN-based method and is close to the current diffusion model. However, the DAE. needs additional training of an aligner for audio features and image features. When we remove the landmark information and directly use the DeepSpeech features as the controllable information (replacing the text prompt), the performance of our model ($Ours^1$) significantly deteriorates. This underscores the efficiency of the landmark as additional information.

\begin{table}[htbp]
\small
\center
\caption{The quantitative results of the compared methods}
\begin{tabular}{l|lll|ll}
\hline
      & Wav2Lip               & MakeitT.            & DAE.              & $Ours^1$ & $Ours^2$ \\ \hline
LSE-D &       7.83            &     9.08            &  6.98             &    10.00      &  \textbf{6.82}   \\ \hline
LSE-C &       7.81            &     5.88            &  \textbf{9.25}    &    5.04     &  9.01   \\ \hline
LD   & \multicolumn{1}{c}{-} & \multicolumn{1}{c}{-} & \multicolumn{1}{c|}{-} &   0.102     &  \textbf{0.004}    \\ \hline
PSNR  & \multicolumn{1}{c}{-} & \multicolumn{1}{c}{-} & \multicolumn{1}{c|}{-} &        15.56 &   \textbf{24.31}   \\ \hline
SSIM  & \multicolumn{1}{c}{-} & \multicolumn{1}{c}{-} & \multicolumn{1}{c|}{-} &        0.67 &   \textbf{0.94}   \\ \hline
\end{tabular}
\label{compaire}
\end{table}

\begin{figure}[htbp]
\centering
\includegraphics[width=\linewidth]{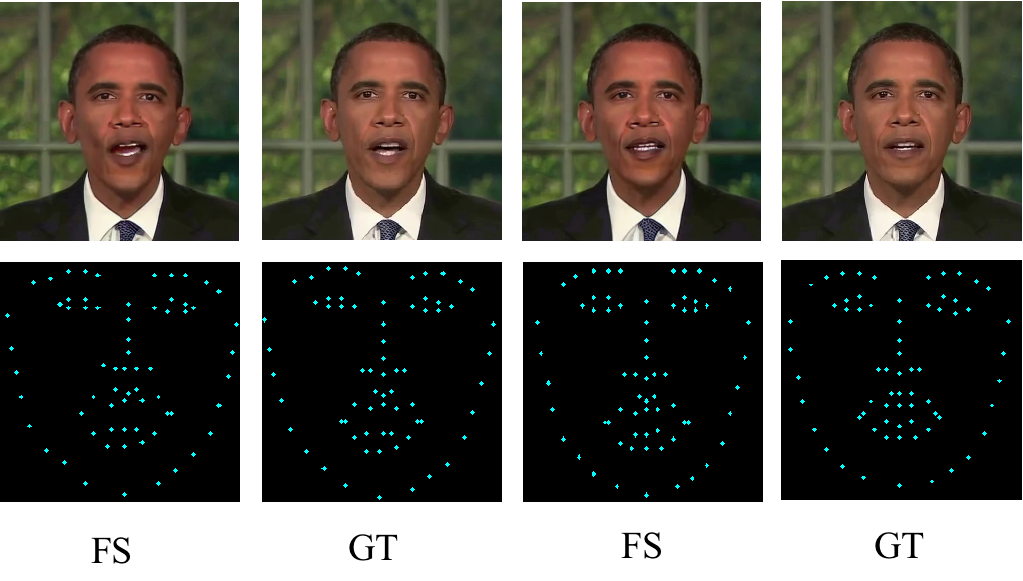}
\caption{The results of the face synthesis (FS) with landmark guidance. The lower part displays the corresponding landmark results.}
\label{face synthesis}
\end{figure}

Figure~\ref{face synthesis} showcases the qualitative outcomes obtained from our face synthesis network. The illustration highlights that the generated faces exhibit a coherent geometric structure across various head poses, owing to the guidance from the upper face and complete landmark information. The lower part of the figure displays the landmark prediction results, reflecting the geometric accuracy of the synthesized face.

\section{conculution}
In this paper, we introduce DiffTalker, a co-driven diffusion method tailored for generating talking faces. DiffTalker consists of two sub-agents: a landmark completion network and a face synthesis network. By harnessing landmarks, we establish a seamless connection between the audio domain and images, thereby enhancing the network's capability to generate precise geometry and mouth shape results. Experimental results demonstrate DiffTalker's exceptional performance in producing clear and geometrically accurate talking faces, all without the requirement for additional alignment between audio and image features. 

\vfill\pagebreak


\bibliographystyle{IEEEbib}
\bibliography{refs}

\begin{thebibliography}{10}

\bibitem{thite2022digital}
Mohan Thite,
\newblock ``Digital human resource development: where are we? where should we
  go and how do we go there?,''
\newblock {\em Human Resource Development International}, vol. 25, no. 1, pp.
  87--103, 2022.

\bibitem{ji2022eamm}
Xinya Ji, Hang Zhou, Kaisiyuan Wang, Qianyi Wu, Wayne Wu, Feng Xu, and Xun Cao,
\newblock ``Eamm: One-shot emotional talking face via audio-based emotion-aware
  motion model,''
\newblock in {\em ACM SIGGRAPH 2022 Conference Proceedings}, 2022, pp. 1--10.

\bibitem{chen2019hierarchical}
Lele Chen, Ross~K Maddox, Zhiyao Duan, and Chenliang Xu,
\newblock ``Hierarchical cross-modal talking face generation with dynamic
  pixel-wise loss,''
\newblock in {\em Proceedings of the IEEE/CVF conference on computer vision and
  pattern recognition}, 2019, pp. 7832--7841.

\bibitem{song2018talking}
Yang Song, Jingwen Zhu, Dawei Li, Xiaolong Wang, and Hairong Qi,
\newblock ``Talking face generation by conditional recurrent adversarial
  network,''
\newblock {\em arXiv preprint arXiv:1804.04786}, 2018.

\bibitem{yin2022styleheat}
Fei Yin, Yong Zhang, Xiaodong Cun, Mingdeng Cao, Yanbo Fan, Xuan Wang, Qingyan
  Bai, Baoyuan Wu, Jue Wang, and Yujiu Yang,
\newblock ``Styleheat: One-shot high-resolution editable talking face
  generation via pre-trained stylegan,''
\newblock in {\em European conference on computer vision}. Springer, 2022, pp.
  85--101.

\bibitem{prajwal2020lip}
KR~Prajwal, Rudrabha Mukhopadhyay, Vinay~P Namboodiri, and CV~Jawahar,
\newblock ``A lip sync expert is all you need for speech to lip generation in
  the wild,''
\newblock in {\em Proceedings of the 28th ACM international conference on
  multimedia}, 2020, pp. 484--492.

\bibitem{guan2023stylesync}
Jiazhi Guan, Zhanwang Zhang, Hang Zhou, Tianshu Hu, Kaisiyuan Wang, Dongliang
  He, Haocheng Feng, Jingtuo Liu, Errui Ding, Ziwei Liu, et~al.,
\newblock ``Stylesync: High-fidelity generalized and personalized lip sync in
  style-based generator,''
\newblock in {\em Proceedings of the IEEE/CVF Conference on Computer Vision and
  Pattern Recognition}, 2023, pp. 1505--1515.

\bibitem{zhou2020makelttalk}
Yang Zhou, Xintong Han, Eli Shechtman, Jose Echevarria, Evangelos Kalogerakis,
  and Dingzeyu Li,
\newblock ``Makelttalk: speaker-aware talking-head animation,''
\newblock {\em ACM Transactions On Graphics (TOG)}, vol. 39, no. 6, pp. 1--15,
  2020.

\bibitem{rombach2022high}
Robin Rombach, Andreas Blattmann, Dominik Lorenz, Patrick Esser, and Bj{\"o}rn
  Ommer,
\newblock ``High-resolution image synthesis with latent diffusion models,''
\newblock in {\em Proceedings of the IEEE/CVF conference on computer vision and
  pattern recognition}, 2022, pp. 10684--10695.

\bibitem{hu2021lora}
Edward~J Hu, Yelong Shen, Phillip Wallis, Zeyuan Allen-Zhu, Yuanzhi Li, Shean
  Wang, Lu~Wang, and Weizhu Chen,
\newblock ``Lora: Low-rank adaptation of large language models,''
\newblock {\em arXiv preprint arXiv:2106.09685}, 2021.

\bibitem{zhang2023adding}
Lvmin Zhang and Maneesh Agrawala,
\newblock ``Adding conditional control to text-to-image diffusion models,''
\newblock {\em arXiv preprint arXiv:2302.05543}, 2023.

\bibitem{du2023dae}
Chenpng Du, Qi~Chen, Tianyu He, Xu~Tan, Xie Chen, Kai Yu, Sheng Zhao, and Jiang
  Bian,
\newblock ``Dae-talker: High fidelity speech-driven talking face generation
  with diffusion autoencoder,''
\newblock {\em arXiv preprint arXiv:2303.17550}, 2023.

\bibitem{zhua2023audio}
Yizhe Zhua, Chunhui Zhanga, Qiong Liub, and Xi~Zhoub,
\newblock ``Audio-driven talking head video generation with diffusion model,''
\newblock in {\em ICASSP 2023-2023 IEEE International Conference on Acoustics,
  Speech and Signal Processing (ICASSP)}. IEEE, 2023, pp. 1--5.

\bibitem{wang2021one}
Ting-Chun Wang, Arun Mallya, and Ming-Yu Liu,
\newblock ``One-shot free-view neural talking-head synthesis for video
  conferencing,''
\newblock in {\em Proceedings of the IEEE/CVF conference on computer vision and
  pattern recognition}, 2021, pp. 10039--10049.

\bibitem{cudeiro2019capture}
Daniel Cudeiro, Timo Bolkart, Cassidy Laidlaw, Anurag Ranjan, and Michael~J
  Black,
\newblock ``Capture, learning, and synthesis of 3d speaking styles,''
\newblock in {\em Proceedings of the IEEE/CVF Conference on Computer Vision and
  Pattern Recognition}, 2019, pp. 10101--10111.

\bibitem{amodei2016deep}
Dario Amodei, Sundaram Ananthanarayanan, Rishita Anubhai, Jingliang Bai, Eric
  Battenberg, Carl Case, Jared Casper, Bryan Catanzaro, Qiang Cheng, Guoliang
  Chen, et~al.,
\newblock ``Deep speech 2: End-to-end speech recognition in english and
  mandarin,''
\newblock in {\em International conference on machine learning}. PMLR, 2016,
  pp. 173--182.

\bibitem{guo2021ad}
Yudong Guo, Keyu Chen, Sen Liang, Yong-Jin Liu, Hujun Bao, and Juyong Zhang,
\newblock ``Ad-nerf: Audio driven neural radiance fields for talking head
  synthesis,''
\newblock in {\em Proceedings of the IEEE/CVF International Conference on
  Computer Vision}, 2021, pp. 5784--5794.

\bibitem{kingma2013auto}
Diederik~P Kingma and Max Welling,
\newblock ``Auto-encoding variational bayes,''
\newblock {\em arXiv preprint arXiv:1312.6114}, 2013.

\bibitem{suwajanakorn2017synthesizing}
Supasorn Suwajanakorn, Steven~M Seitz, and Ira Kemelmacher-Shlizerman,
\newblock ``Synthesizing obama: learning lip sync from audio,''
\newblock {\em ACM Transactions on Graphics (ToG)}, vol. 36, no. 4, pp. 1--13,
  2017.

\end{thebibliography}

\end{document}